\documentclass{article}
\usepackage[nonatbib, preprint]{neurips_2024}

\usepackage[utf8]{inputenc} 
\usepackage[T1]{fontenc}    
\usepackage{hyperref}       
\usepackage{url}            
\usepackage{booktabs}       
\usepackage{amsfonts}       
\usepackage{nicefrac}       
\usepackage{microtype}      
\usepackage{xcolor}         
\usepackage{amsmath}
\usepackage{bm}
\usepackage{algorithm}
\usepackage{algpseudocode}
\usepackage{graphicx} 
\usepackage{caption}
\usepackage{subfigure}
\usepackage{subcaption}
\graphicspath{{./images/}} 
\usepackage[backend=biber,style=numeric]{biblatex}
\title{ADBA:Approximation Decision Boundary Approach for Black-Box Adversarial Attacks}

\author{%
  Feiyang Wang, Xingquan Zuo, Hai Huang \\
  Department of Computer Science and Technology\\
  Beijing University of Post and Telecommunication, Beijing, China\\
  \texttt{\{2016211185, zuoxq, hhuang\}@bupt.edu.cn} \\
  \And
  Gang Chen \\
  School of Engineering and Computer Science\\
  Victoria University of Wellington, Wellington, New Zealand \\
  \texttt{aaron.chen@vuw.ac.nz}
}

\begin{document}

\maketitle

\begin{abstract}
\label{abstract}

Many machine learning models are susceptible to adversarial attacks, with decision-based black-box attacks representing the most critical threat in real-world applications. These attacks are extremely stealthy, generating adversarial examples using hard labels obtained from the target machine learning model. This is typically realized by optimizing \emph{perturbation directions}, guided by decision boundaries identified through query-intensive exact search, significantly limiting the attack success rate. This paper introduces a novel approach using the Approximation Decision Boundary (ADB) to efficiently and accurately compare perturbation directions without precisely determining decision boundaries. The effectiveness of our ADB approach (ADBA) hinges on promptly identifying suitable ADB, ensuring reliable differentiation of all perturbation directions. For this purpose, we analyze the probability distribution of decision boundaries, confirming that using the distribution's median value as ADB can effectively distinguish different perturbation directions, giving rise to the development of the ADBA-md algorithm. ADBA-md only requires four queries on average to differentiate any pair of perturbation directions, which is highly query-efficient. Extensive experiments on six well-known image classifiers clearly demonstrate the superiority of ADBA and ADBA-md over multiple state-of-the-art black-box attacks. The source code is available at \url{https://github.com/BUPTAIOC/ADBA}.

\end{abstract}

\vspace{-0.2cm}
\section{Introduction}
\vspace{-0.2cm}
\label{section Introduction}

{\bf Background and motivation}. It is widely acknowledged that machine learning methods such as \emph{Deep Neural Networks} (DNNs) are vulnerable to adversarial attacks \cite{xie_feature_2019, xie_adversarial_2017}. Particularly, for image classification tasks, tiny additive perturbations in the input images can significantly affect the classification accuracy of a pre-trained model \cite{xie_feature_2019}. The impact of these intentionally designed perturbations in real-world scenarios \cite{eykholt_robust_2018,lin_sensitive_2023} has heightened security worries for critical applications of deep neural networks in many domains \cite{eykholt_robust_2018,li_towards_2023, li_physical-world_2023,li2020deep}, which is detailed in Appendix \ref{appendix societal impacts}.

Adversarial attacks can be categorized into \emph{white-box} attacks and \emph{black-box} attacks \cite{long_survey_2022}. White-box attacks \cite{wang_enhancing_2021} require the attacker to have comprehensive knowledge of the target machine learning model, rendering them impractical in many real-world scenarios. In comparison, black-box attacks \cite{bai_query_2023,brendel_decision-based_2018,li_aha_2021} are more realistic since they do not required detailed knowledge of the target model.

Black-box attacks can be divided into \emph{transfer-based attacks}, \emph{score-based attacks} (or \emph{soft-label attacks, gray-box attacks}), and \emph{decision-based attacks} (\emph{hard-label attacks})  \cite{li_aha_2021}. More detailed discussion of these attacks can be found in Appendix~\ref{appendix Related work}. Among them, decision-based attacks \cite{brendel_decision-based_2018,li_aha_2021,shi_decision-based_2022,chen_hopskipjumpattack_2020,cheng_sign-opt_2020} are extremely stealthy since they rely solely on the hard label from the target model to create adversarial examples.

This paper studies the decision-based attacks due to their general applicability and effectiveness in real-world adversarial situations. These attacks aim to deceive the target model while adhering to two constraints \cite{ilyas_black-box_2018}: 1) they must generate adversarial examples with as few queries as possible (i.e., query-efficient) and cannot exceed a predetermined number of queries (i.e., {\bf query budget}), and 2) the strength of the adversarial perturbations must remain within a predefined threshold \( \epsilon \). Violating those constrains results in the attack being easily detected or deemed unsuccessful. These constraints bring huge challenges to attackers \cite{ilyas_black-box_2018}. Specifically, lacking detailed knowledge of the target model and its output scores poses tremendous difficulty for attackers to determine the \emph{decision boundary} (i.e., the minimum perturbation strength required to deceive the model) with respect to any perturbation direction \cite{cheng_sign-opt_2020}. Thus, decision-based attacks often require a large number of queries to identify the decision boundary and optimize the perturbation direction, increasing the likelihood of being detected and hurting the attack success rate. Therefore, enhancing attack efficiency and minimizing the number of queries are essential for decision-based attacks \cite{bai_query_2023,chen_hopskipjumpattack_2020,cheng_sign-opt_2020,ilyas_black-box_2018}.

\begin{figure}[t]
\centering 
    \subfigure[Too large ADB fails to compare \(g(\bm{d_1}), g(\bm{d_2})\)]{
        \label{Fig1.sub.1}
        \includegraphics[width=0.32\textwidth,height = 4.5cm]{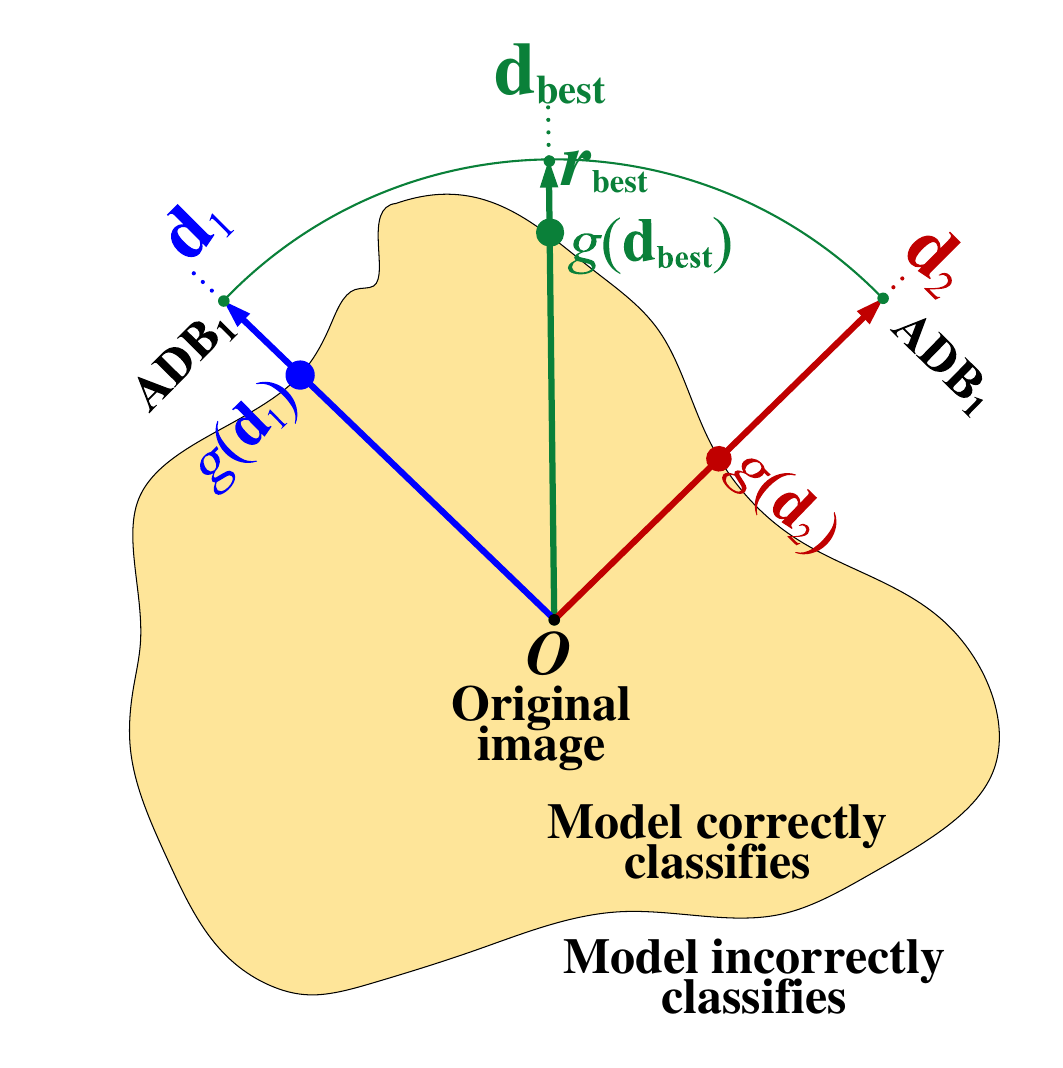}}
    \subfigure[Too small ADB fails to compare \(g(\bm{d_1}), g(\bm{d_2})\)]{
        \label{Fig1.sub.2}
        \includegraphics[width=0.32\textwidth,height = 4.5cm]{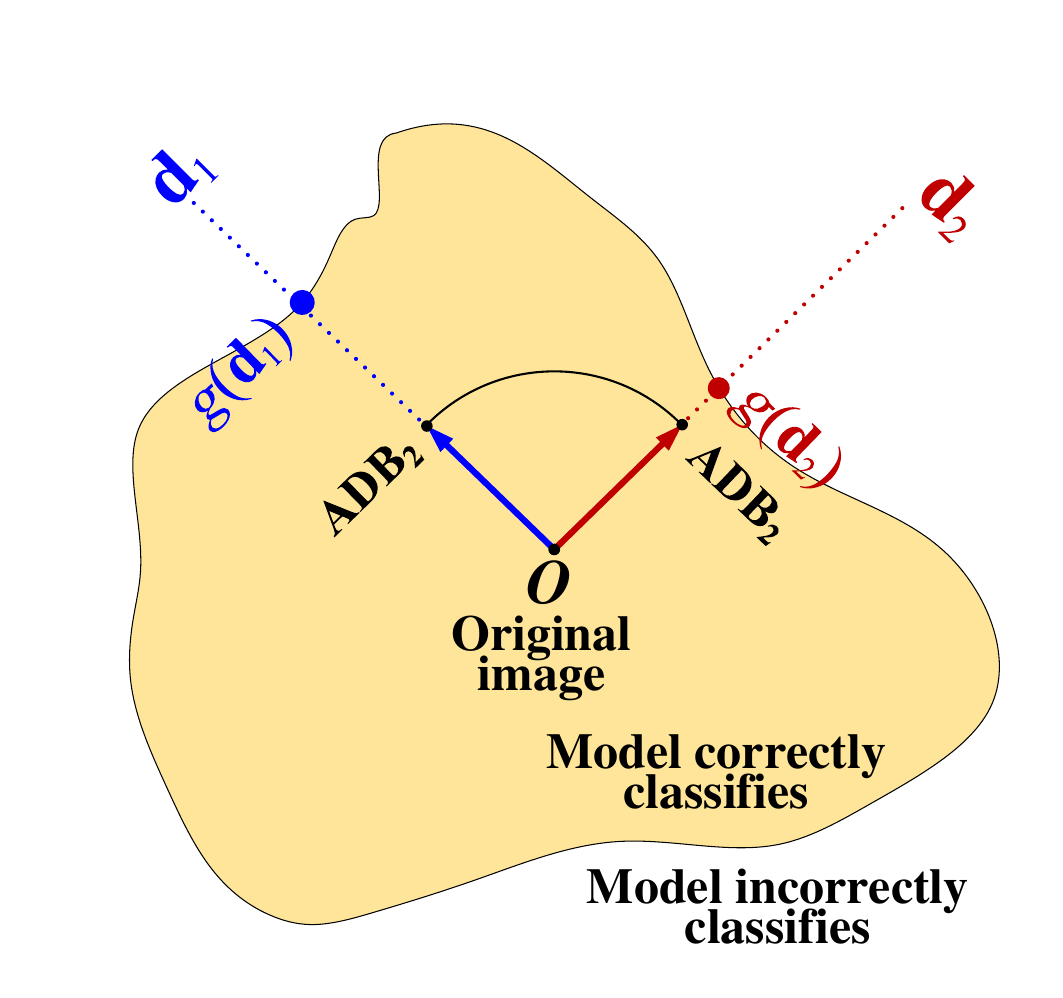}}
    \subfigure[ADB succeed in comparing \(g(\bm{d_1}), g(\bm{d_2})\)]{
        \label{Fig1.sub.3}
        \includegraphics[width=0.32\textwidth,height = 4.5cm]{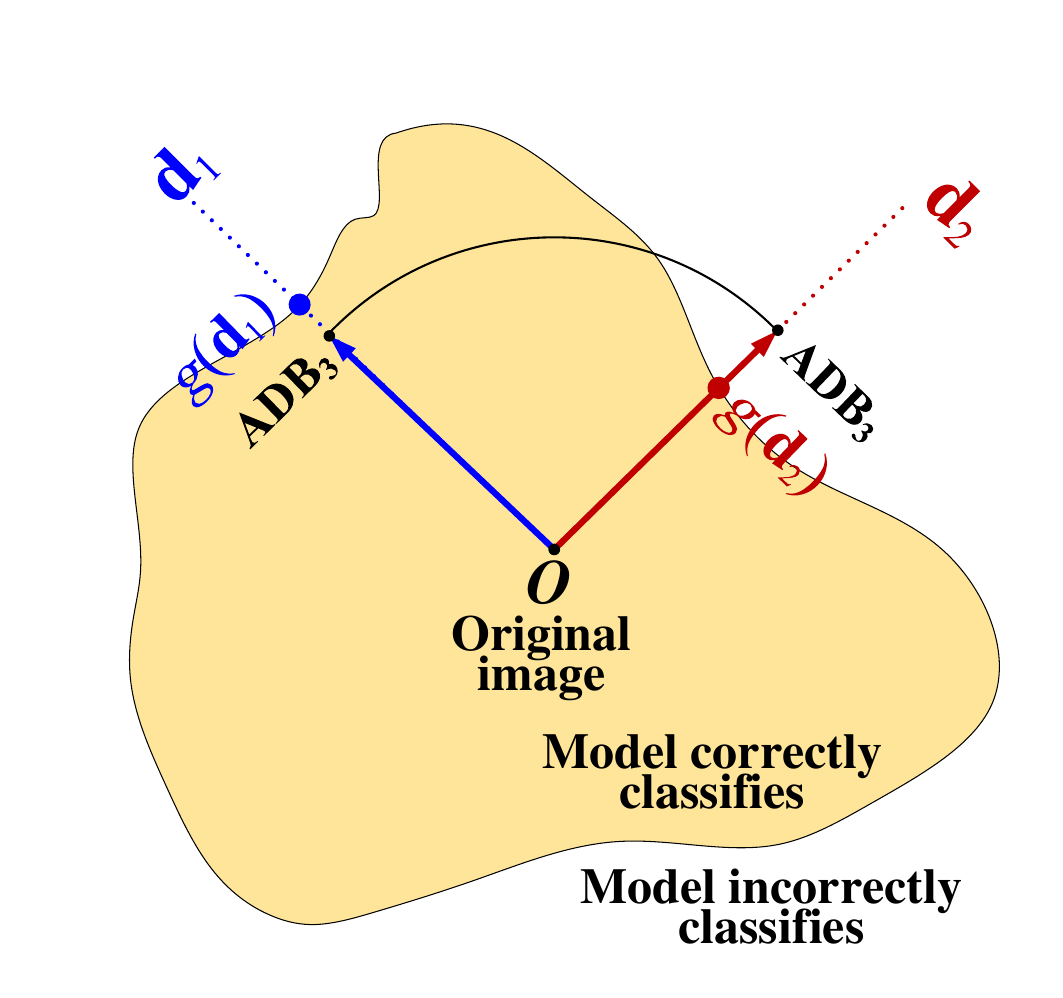}}
    \caption{Compare directions \( \bm{d_1} \) and \( \bm{d_2} \) using ADB. In (a), comparison of \( \bm{d_1} \) and \( \bm{d_2} \) fails because \( \mathrm{\text{ADB}_1} \) is too large (\( g(\bm{d_1}) \leq \mathrm{\text{ADB}_1} \) and \( g(\bm{d_2}) \leq \mathrm{\text{ADB}_1} \)). In (b), comparison of \( \bm{d_1} \) and \( \bm{d_2} \) fails because \( \mathrm{\text{ADB}_2} \) is too small (\( g(\bm{d_1}) > \mathrm{\text{ADB}_2} \) and \( g(\bm{d_2}) > \mathrm{\text{ADB}_2} \)). In (c), \(\bm{d_2}\) is superior to \(\bm{d_1}\) and \(g(\bm{d_2}) \leq \text{ADB}_3 < g(\bm{d_1})\) because the attack in direction \(\bm{d_1}\) fails but the attack in direction \(\bm{d_1}\) succeeds. \(\text{ADB}_3\) is hence the ADB of \(\bm{d_2}.\)}
    \label{1}
\end{figure}

{\bf Proposed method}. Existing decision-based attacks can be divided into \emph{random search attacks} \cite{brendel_decision-based_2018,li_aha_2021,cheng_sign-opt_2020,brunner_guessing_2019,chen_rays_2020,cheng_query-efficient_2018}(reviewed in detail in Section~\ref{section Related_work}), \emph{gradient estimation attacks} \cite{chen_hopskipjumpattack_2020,li_qeba_2020,liu_geometry-inspired_2019,rahmati_geoda_2020}, and \emph{geometric modeling attacks} \cite{avidan_triangle_2022,reza_cgba_2023,maho_surfree_2021}(reviewed in detail in Appendix~\ref{appendix Related work}). 
This paper focuses on random search attacks, aiming to find the optimal perturbation direction with the smallest decision boundary. 
For this purpose, query-intensive exact search techniques such as binary search are typically utilized to identify the decision boundaries of different perturbation directions~\cite{cheng_sign-opt_2020, chen_rays_2020}. 
For two candidate perturbation directions \(\bm{d_1}\) and \(\bm{d_2}\). Binary search is used to calculate their decision boundaries \(g(\bm{d_1})\) and \(g(\bm{d_2})\), and determine which direction has the smaller decision boundary~\cite{chen_rays_2020}. However, binary search demands for a large number of queries, resulting in poor query efficiency.

In this study, we show that different perturbation directions can be compared without knowing their precise decision boundaries, avoiding costly exact search methods, as demonstrated in Figure~\ref{Fig1.sub.3}. Suppose that an approximation decision boundary (ADB) enables direction \(\bm{d_2}\) to deceive the target model but direction \(\bm{d_1}\) fails to do the same, we can conclude that \(g(\bm{d_2}) \leq \text{ADB} < g(\bm{d_1})\) and \(\bm{d_2}\) is deemed superior for the black-box attack. Here, \(g(\bm{d_1}) \text{and } g(\bm{d_2})\) refer to the actual decision boundaries of \(\bm{d_1} \text{and } \bm{d_2}\). On the contrary, \(g(\bm{d_1})\) and \(g(\bm{d_2})\) cannot be clearly differentiated upon using ADB that is either too large or too small, as shown in Figure~\ref{Fig1.sub.1} and Figure~\ref{Fig1.sub.2}. Hence, a key issue is to quickly find an appropriate ADB with few queries. To tackle this issue, we analyze the probability distribution of decision boundaries and discover that it is highly likely for any pair of perturbation directions to be successfully differentiated upon using this distribution's median value as ADB. In fact, only four queries on average are required for this purpose. Based on this idea, we propose innovative \emph{Approximation Decision Boundary Approach} (ADBA) and \emph{Median-Search Based ADBA} (ADBA-md) for random search attacks.

{\bf Comprehensive experiments}. We perform comprehensive experiments for decision-based black-box attacks using a variety of image classification models. We select 6 well-known deep models that span a wide range of architectures. We compare our methods with four state-of-the-art decision-based adversarial attack approaches. The results show that our methods can generate adversarial examples with a high attack success rate (i.e., fooling rate) while using a small number of queries.

{\bf This paper makes three main contributions}. (1) We propose a novel Approximation Decision Boundary Approach (ADBA) for query-efficient decision-based attacks. (2) We analyze the distribution of decision boundaries and discover that using the distribution's median value as ADB can differentiate any pair of perturbation directions with high query efficiency, giving rise to the development of ADBA-md. (3) We conduct comprehensive experiments, demonstrating that ADBA and ADBA-md can significantly outperform four leading decision-based attack approaches across six modern deep models on the ImageNet dataset.

\vspace{-0.2cm}
\section{Related work}
\vspace{-0.2cm}
\label{section Related_work}
Random search attacks adopt a random search framework to generate candidate perturbation directions and adjust them according to the sizes of decision boundaries.  \cite{brendel_decision-based_2018,brunner_guessing_2019,cheng_query-efficient_2018,cheng_sign-opt_2020,chen_rays_2020}. These boundaries are precisely identified through methods such as binary search, which requires a large number of queries. For example, Boundary Attack \cite{brendel_decision-based_2018}, Biased Boundary Attack \cite{brunner_guessing_2019}, and AHA \cite{li_aha_2021} progressively alter the current perturbation direction through a random walk along the decision boundary, which is query-intensive. To reduce the number of queries in \cite{brendel_decision-based_2018}, Biased Boundary Attack \cite{brunner_guessing_2019} proposes a biased sampling framework to accelerate the search for perturbation directions. Meanwhile, AHA in \cite{li_aha_2021} gathers information from all historical queries as a prior for current sampling decisions to improve the efficiency of random walk. Additionally, OPT and Sign-OPT developed respectively in \cite{cheng_query-efficient_2018} and \cite{cheng_sign-opt_2020} transform the hard-label black-box attack to a continuous optimization problem, solvable through zeroth-order optimization. However, they utilize binary search to identify the decision boundary of each perturbation direction, demanding for a large number of queries. RayS in \cite{chen_rays_2020} is a state-of-the-art decision-based attack based on the \(l_\infty\) norm. It transforms the task of optimizing the perturbation direction from the continuous domain to the discrete domain. It generates perturbation directions by progressively dividing and inverting direction vectors, with each division resulting in smaller blocks as the optimization advances, noticeably enhancing the efficiency in finding good directions. However, RayS optimizes perturbation directions based on precise estimation of the decision boundary through a query-intensive binary search process. More discussion of related works, including gradient estimation attacks and geometric modeling attacks, can be found in Appendix~\ref{appendix Related work}.

\vspace{-0.2cm}
\section{Proposed Method}
\vspace{-0.2cm}
\label{section Proposed_Method}

\subsection{Preliminaries}
\label{section Preliminaries}
Let \( F\colon \mathbb{X}^N \rightarrow \{1, \dots, K\} \) denote a target image classification model that assigns images to one of \( K \) distinct classes. The model’s input is a normalized RGB image \( \bm{x} \in \mathbb{X}^N \), where \( N = \text{Width} \times \text{Height} \times \text{Channels} \) represents the dimensionality of the image, encompassing all pixels across all channels. In this context, \( \mathbb{X}^N \) is the input space of all images to be classified. The channel value of each pixel of any image \(\bm{x}\in \mathbb{X}^N\) ranges between 0 and 1. Meanwhile, \( y(\bm{x}) \in \{1, \dots, K\} \) denotes the true label of image \( \bm{x} \), and \( F(\bm{x}) \) is the label produced by the classification model \( F \) for image \( \bm{x} \). The goal of an adversarial attack is to find an adversarial example \( \tilde{\bm{x}} \in \mathbb{X}^N \) such that \( F(\tilde{\bm{x}}) \neq y(\bm{x}) \) (untargeted attack) or \( F(\tilde{\bm{x}}) = t \) (targeted attack, where \(t\) is a given target label, and \( t \neq y(\bm{x}) \)), subject to the condition that \( \|\tilde{\bm{x}} - \bm{x}\|_v \leq \epsilon \), where \( \bm{x} \) is a correctly classified image. Here, \( \epsilon \) refers to the allowed maximum perturbation strength and \( v \) refers to the norm used to measure the perturbation strength, such as \( l_1 \), \( l_2 \), and \( l_\infty \) norms \cite{long_survey_2022}. In this paper, we adopt the \( l_\infty \) norm, following RayS \cite{chen_rays_2020}. This paper focuses primarily on untargeted attacks, aiming to force the target image classification models to produce arbitrary incorrect class labels for any image \(\bm{x}\). This goal can be expressed as follows:
\begin{equation}
\label{equation_attack_question_original}
\max_{\tilde{\bm{x}}} f(\bm{x}, y(\bm{x}), \tilde{\bm{x}}) \quad \text{subject to} \quad \|\tilde{\bm{x}} - \bm{x}\|_{\infty} \leq \epsilon \tag{1}
\end{equation}
\[
\text{where} \quad f(\bm{x}, y(\bm{x}), \tilde{\bm{x}}) = 
\begin{cases} 
0 & \text{if } F(\tilde{\bm{x}}) \neq y(\bm{x}) \\
-1 & \text{if } F(\tilde{\bm{x}}) = y(\bm{x})
\end{cases}
\]
Let the adversarial perturbation \(\tilde{\bm{x}} - \bm{x} = r \cdot \bm{d}\), where \(\bm{d} \in [-1,1]^N\) represents the perturbation direction and \(r \in [0, \epsilon]\) represents the perturbation strength. Prior works \cite{ilyas_black-box_2018, chen_rays_2020, cheng_query-efficient_2018,moon_parsimonious_2019,chen_frank-wolfe_2020}, proved that the optimal solution to question (\ref{equation_attack_question_original}) is at the extremities of the \(\ell_{\infty}\) norm ball, i.e., \(\bm{d} \in \{-1, 1\}^N\). In this case, \(\|\tilde{\bm{x}} - \bm{x}\|_{\infty} = \|r \cdot \bm{d}\|_{\infty} = r \cdot \|\bm{d}\|_{\infty} = r\). Hence, the continuous optimization problem in question (\ref{equation_attack_question_original}) is transformed into a mixed-integer optimization problem according to \cite{chen_rays_2020}:
\[
\label{equation_attack_question_01}
\max_{r \in [0,\epsilon], \bm{d} \in \{-1, 1\}^N} h(\bm{x}, y(\bm{x}), r, \bm{d}) \quad  \tag{2}\]
\[\text{where} \quad h(\bm{x}, y(\bm{x}), r, \bm{d}) = 
\begin{cases} 
0 & \text{if } F(\bm{x} + r \cdot \bm{d}) \neq y(\bm{x}) \\
-1 & \text{if } F(\bm{x} + r \cdot \bm{d}) = y(\bm{x})
\end{cases}\]
Previous works \cite{cheng_sign-opt_2020,ilyas_black-box_2018,chen_rays_2020,cheng_query-efficient_2018} also proved that the solution space of question (\ref{equation_attack_question_01}) is continuous, and  exploited the concept of the decision boundary to develop effective adversarial attacks. For any direction \(\bm{d}\), there exists a \emph{decision boundary} \(g(\bm{d})\) such that \(F(\bm{x} + g(\bm{d}) \cdot \bm{d}) \neq y(\bm{x})\) and \(F(\bm{x} + (g(\bm{d}) - \Delta) \cdot \bm{d}) = y(\bm{x})\), \( \forall \Delta\in (0, \tau]\). \(g(\bm{d})\) is further defined in question (\ref{equation_attack_question_rays}) below. As shown in Figure~\ref{Fig1.sub.1},  \(g(\bm{d_{\text{best}}})\), \(g(\bm{d_1})\), and \(g(\bm{d_2})\) are the decision boundaries of the previous best direction \(\bm{d_{\text{best}}}\) and two new candidate directions \(\bm{d_1}\), \(\bm{d_2}\), respectively. According to the definition of current decision boundary in \cite{chen_rays_2020}, the decision-based attacks have the goal to find the optimal perturbation direction \(\bm{d_{\text{best}}}\): 
\[
\label{equation_attack_question_rays}
\bm{d_{\text{best}}}=\operatorname*{argmin}_{\bm{d} \in \{-1,1\}^N} g(\bm{d}) \quad \text{with} \quad g(\bm{d}) = \inf \{F(\bm{x} + r \cdot \bm{d}) \neq y(\bm{x}) \} \tag{3}\]
According to the above equation, the direction \(\bm{d_2}\) in Figure~\ref{Fig1.sub.1} has the smallest decision boundary \(g(\bm{d_2})\) among \(g(\bm{d_{\text{best}}})\), \(g(\bm{d_1})\), and \(g(\bm{d_2})\).

\begin{algorithm}[htbp]
\caption{Approximate Decision Boundary Approach}
\label{algorithm_ADBA}
\textbf{Input}: Model \(F\), the original image \(\bm{x}\) and its label \(y(\bm{x})\), and maximum perturbation strength \(\epsilon\);\\
\textbf{Output}: Optimal direction \(\bm{d_{\text{best}}}\) with approximation decision boundary \(r_{\text{best}}\);\\
\textbf{Initialization}: Initialize current best direction \(\bm{d_{\text{best}}} \leftarrow (1,\ldots,1)\), and set current best perturbation strength \(r_{\text{best}} \leftarrow 1\), block variable \(\leftarrow 0\) and block index \(k \leftarrow 0\);
\begin{algorithmic}[1]
\While{remaining query budget \(> 0\) and \(r_{\text{best}} > \epsilon\)}
\State \(L = N/2^{(s+1)}\);
\State \(\bm{d_1} \leftarrow \bm{d_{\text{best}}}\), \(\bm{d_2} \leftarrow \bm{d_{\text{best}}}\);
\For{\(i \in [k \cdot L, (k+1) \cdot L)\)}
\State \(\bm{d_1}[i] \leftarrow -\bm{d_1}[i]\)
\EndFor
\For{\(i \in [(k+1) \cdot L, (k+2) \cdot L)\)}
\State \(\bm{d_2}[i] \leftarrow -\bm{d_2}[i]\)
\EndFor
\State \(\bm{d_{\text{best}}}, r_{\text{best}} \leftarrow \text{Algorithm 2}(F, \{\bm{x},y(\bm{x})\}, \bm{d_{\text{best}}}, r_{\text{best}}, \bm{d_1}, \bm{d_2})\);
\State \(k \leftarrow k + 2\);
\If{\(k = 2^{(s+1)}\)}
\State \(s \leftarrow s + 1\)
\State \(k \leftarrow 0\)
\EndIf
\EndWhile
\State \Return \(\bm{d_{\text{best}}}, r_{\text{best}}\)
\end{algorithmic}
\end{algorithm}

\subsection{Approximation Decision Boundary Approach (ADBA)}
\label{section ADBA}
To compare the decision boundary \(g(\bm{d})\) of different directions, existing methods typically employ exact search techniques such as binary search \cite{cheng_sign-opt_2020}\cite{ilyas_black-box_2018} \cite{chen_rays_2020} to identify \(g(\bm{d})\) with high accuracy (e.g., up to 0.001 \cite{chen_rays_2020}). Obviously, this process requires a large number of queries. We notice that, in the early stages of optimization, directions are often far away from optima and their decision boundaries often exceed the threshold \(\epsilon\) on the perturbation strength. Thus it is unnecessary to precisely identify the decision boundary. However, existing works \cite{cheng_query-efficient_2018, cheng_sign-opt_2020, chen_rays_2020} perform numerous queries to determine the accurate decision boundary, even in the early stages, resulting in poor query efficiency. Different from these works, in this paper, we demonstrate that perturbation directions can be reliably compared and optimized without knowing the precise decision boundary. Driven by this idea, we propose ADBA, as summarized in Algorithm~\ref{algorithm_ADBA}.

In Algorithm~\ref{algorithm_ADBA}, ADBA iteractively searches for the optimal perturbation direction. It keeps track of the current best direction  \( \bm{d_{\text{best}}} \) along with its ADB \( r_{\text{best}} \) , and update \( \bm{d_{\text{best}}} \) and \( r_{\text{best}} \) in each iteration. The initial perturbation direction \( \bm{d_{\text{best}}} \) is flattened into a one-dimensional vector \((1,\ldots,1)\) with \(N=\text{Width} \times \text{Height} \times \text{Channels}\) dimensions. The current best perturbation strength \( r_{\text{best}} \) that represents the ADB of \( \bm{d_{\text{best}}} \) is set to 1 initially to make sure that \(F(\bm{x} + r_{\text{best}} \cdot \bm{d_{\text{best}}}) \neq y(\bm{x})\). In line with RayS \cite{chen_rays_2020}, we use the block variable \( s \) to control the block size. In line 2, the current best direction vector \( \bm{d_{\text{best}}} \) is divided into \( 2^{(s+1)} \) smaller blocks. In lines 3-9 of each algorithm iteration, we choose two blocks of \( \bm{d_{\text{best}}} \) in sequence and reverse the sign of each block respectively to create two new directions, \( \bm{d_1} \) and \( \bm{d_2} \). In line 10, different from \cite{chen_rays_2020} that uses binary search to accurately identify \(g(\bm{d_1})\) and \(g(\bm{d_2})\), we devise Algorithm~\ref{algorithm_compareADB} (to be detailed in Subsection~\ref{section Compare d using ADB}) to update the optimal direction \( \bm{d_{\text{best}}} \) based on its approximation decision boundary  \( r_{\text{best}} \). In lines 11-15, if the block index \( k \) reaches \( 2^{(s+1)} \), i.e., all current blocks have been utilized for reversal, then \( 2^{(s+2)} \) more directions are produced in subsequent iterations by \( s \leftarrow s+1 \). If the  ADB \( r_{\text{best}} \) of \( \bm{d_{\text{best}}} \) is within the allowed strength \( \epsilon \) or the number of queries exceeds the budget, Algorithm~\ref{algorithm_ADBA} returns the current best direction together with its perturbation strength. The former scenario indicates a successful attack, while the latter signifies a failed attack.

\begin{algorithm}[htbp]
\caption{Compare Directions Using ADB}
\label{algorithm_compareADB}
\textbf{Input}: Model \(F\), original image \(\bm{x}\) and its label \(y(\bm{x})\), current best direction \(\bm{d_{\text{best}}}\) with approximation decision boundary \(r_{\text{best}}\), two new directions \(\bm{d_1}\), \(\bm{d_2}\), and search tolerance \(\tau\);\\
\textbf{Output}: New best direction \(\bm{d_{\text{best}}}\) with approximation decision boundary \(r_{\text{best}}\)\\
\textbf{Initialization}: Set \(\text{ADB} \leftarrow r_{\text{best}}\), \textit{start} \(\leftarrow 0\), \textit{end} \(\leftarrow r_{\text{best}}\)
\begin{algorithmic}[1]
\If{\(F(\bm{x} + r_{\text{best}} \cdot \bm{d_1}) = y(\bm{x})\) and \(F(\bm{x} + r_{\text{best}} \cdot \bm{d_2}) = y(\bm{x})\)}
\State \Return \(\bm{d_{\text{best}}}, r_{\text{best}}\)
\EndIf
\While{\textit{end} - \textit{start} \( > \tau\)}
\If{\(F(\bm{x} + \text{ADB} \cdot \bm{d_1}) \neq y(\bm{x})\) and \(F(\bm{x} + \text{ADB} \cdot \bm{d_2}) \neq y(\bm{x})\)}
\State \(\text{ADB} \leftarrow (\textit{start} + \textit{end})/2\)
\State \textit{end} \(\leftarrow \text{ADB}\)
\ElsIf{\(F(\bm{x} + \text{ADB} \cdot \bm{d_1}) = y(\bm{x})\) and \(F(\bm{x} + \text{ADB} \cdot \bm{d_2}) = y(\bm{x})\)}
\State \(\text{ADB} \leftarrow (\textit{start} + \textit{end})/2\)
\State \textit{start} \(\leftarrow \text{ADB}\)
\ElsIf{\(F(\bm{x} + \text{ADB} \cdot \bm{d_1}) \neq y(\bm{x})\) and \(F(\bm{x} + \text{ADB} \cdot \bm{d_2}) = y(\bm{x})\)}
\State \Return \(\bm{d_1}, \text{ADB}\)
\ElsIf{\(F(\bm{x} + \text{ADB} \cdot \bm{d_1}) = y(\bm{x})\) and \(F(\bm{x} + \text{ADB} \cdot \bm{d_2}) \neq y(\bm{x})\)}
\State \Return \(\bm{d_2}, \text{ADB}\)
\EndIf
\EndWhile
\State \Return \(\bm{d_1}, \text{ADB}\)
\end{algorithmic}
\end{algorithm}

\subsubsection{Comparing Perturbation Directions Using ADB}
\label{section Compare d using ADB}

In this section, we propose a query-efficient method by comparing perturbation directions using ADB, rather than using binary search in \cite{chen_rays_2020}. Our idea is described in more details below. First, we determine whether \(\bm{d_1}\) and \(\bm{d_2}\) are superior to \(\bm{d_{\text{best}}}\); if not, \(\bm{d_{\text{best}}}\) remains the current best, and there is no need to compare \(\bm{d_1}\) and \(\bm{d_2}\). We denote the ADB \(r_{\text{best}}\) of \(\bm{d_{\text{best}}}\) as \(\text{ADB}_1\) and check whether both \(F(\bm{x} + \text{ADB}_1 \cdot \bm{d_1})\) and \(F(\bm{x} + \text{ADB}_1 \cdot \bm{d_2})\) produce wrong class labels (attacks are successful). If \(\text{ADB}_1\) is too large and both attacks succeed, as shown in Figure~\ref{Fig1.sub.1}, then \(g(\bm{d_1}) \leq \text{ADB}_1\) and \(g(\bm{d_2}) \leq \text{ADB}_1\). In this case, we cannot differentiate \(g(\bm{d_1})\) and \(g(\bm{d_2})\). Hence, \(\text{ADB}\) is decreased to \(\text{ADB}_2 = (0 + \text{ADB}_1)/2\). Subsequently, we check whether \(F(\bm{x} + \text{ADB}_2 \cdot \bm{d_1})\) and \(F(\bm{x} + \text{ADB}_2 \cdot \bm{d_2})\) can lead to incorrect classification. If \(\text{ADB}_2\) is too small and both attacks fail, as demonstrated in Figure~\ref{Fig1.sub.2}, \(g(\bm{d_1}) > \text{ADB}_2\) and \(g(\bm{d_2}) > \text{ADB}_2\). In this case, we still cannot differentiate \(g(\bm{d_1})\) and \(g(\bm{d_2})\). We instead increase ADB to \(\text{ADB}_3 = (\text{ADB}_2 + \text{ADB}_1)/2\), and check whether \(F(\bm{x} + \text{ADB}_3 \cdot \bm{d_1})\) and \(F(\bm{x} + \text{ADB}_3 \cdot \bm{d_2})\) can induce wrong classification. This time, in Figure~\ref{Fig1.sub.3}, the attack along the direction \(\bm{d_2}\) succeeds, and the attack along the direction \(\bm{d_1}\) fails, indicating \(g(\bm{d_2}) \leq \text{ADB}_3 < g(\bm{d_1})\); i.e., \(\bm{d_2}\) is superior to \(\bm{d_1}\). Subsequently, we update \(\bm{d_{\text{best}}} = \bm{d_2}\), \(r_{\text{best}} = \text{ADB}_3\).

The above process is summarized in Algorithm~\ref{algorithm_compareADB}. Notably, for the current best direction \(\bm{d_{\text{best}}}\) in Figure~\ref{Fig1.sub.1}, because its real decision boundary \(g(\bm{d_{\text{best}}}) \leq r_{\text{best}}\), we cannot compare \(g(\bm{d_{\text{best}}})\) and \(g(\bm{d_2})\). However, further comparison of \(\bm{d_{\text{best}}}\) and \(\bm{d_2}\) is unnecessary, since this requires twice as many queries with minimal improvement of  \(g(\bm{d_{\text{best}}})\), as evidenced by our preliminary experiments in Appendix~\ref{appendix ADBA-CCM}.

In Algorithm~\ref{algorithm_compareADB}, ADB is set initially to the ADB of the current best direction \(\bm{d_{\text{best}}}\). In lines 1 and 2, if \(g(\bm{d_{\text{best}}}) \leq r_{\text{best}} < g(\bm{d_1})\) and \(g(\bm{d_{\text{best}}}) \leq r_{\text{best}} < g(\bm{d_2})\) (i.e., \(\bm{d_1}\) and \(\bm{d_2}\) are worse than \(\bm{d_{\text{best}}}\)), it is worthless to perform further comparisons. Algorithm~\ref{algorithm_compareADB} hence returns \(\bm{d_{\text{best}}}\) as its output. In lines 5 to 10, if the decision boundaries of \(\bm{d_1}\) and \(\bm{d_2}\) are both either smaller or larger than ADB, then Algorithm~\ref{algorithm_compareADB} updates the \textit{start} or \textit{end} to narrow the search range for ADB. In lines 11 to 14, if the current ADB leads to a successful attack in one direction but not the other direction, Algorithm~\ref{algorithm_compareADB} reports the successful direction and the corresponding ADB. In line 17, if the search range \textit{end} - \textit{start} is less than a search tolerance threshold \(\tau\), indicating that \(g(\bm{d_1})\) and \(g(\bm{d_2})\) are closely matched, the algorithm returns \(\bm{d_1}\) along with the current ADB. Meanwhile, according to Appendix~\ref{appendix ADBA-CCM}, it is unnecessary to compare \(\bm{d_1}\) with \(\bm{d_{\text{best}}}\) because this comparison requires an excessive number of queries while yielding minimal improvement to \(g(\bm{d_{\text{best}}})\).

\subsubsection{Optimization of ADB}
\label{section optimize ADB}

The practical efficiency of Algorithm~\ref{algorithm_compareADB} depends on the swift identification of ADB that can differentiate any two directions \(\bm{d_1}\) and \(\bm{d_2}\) accurately. In lines 6 and 9 of Algorithm~\ref{algorithm_compareADB}, if the current ADB fails to differentiate the two directions, the next search point is chosen as the middle point of the search range \([start, end]\), i.e., ADB \(\leftarrow\) \((start + end)/2\). However, this simple method cannot achieve the desired query efficiency. In this paper, by considering the statistical distribution of decision boundaries, we can identify a more suitable ADB to improve the likelihood of differentiating \(\bm{d_1}\) and \(\bm{d_2}\). Based on this idea, we propose ADBA-md in this section.

We define the random events \( A \), \( B \), \( C \) and \( D \) as follows: 
\begin{align*}
\label{equation_opt_ADB_eventsABCD}
\begin{cases}
A &= \{F(\bm{x} + \text{ADB} \cdot \bm{d_1}) \neq y(\bm{x})\} = \{g(\bm{d_1}) \leq \text{ADB}\}\\
B &= \{F(\bm{x} + \text{ADB} \cdot \bm{d_2}) \neq y(\bm{x})\} = \{g(\bm{d_2}) \leq \text{ADB}\}\\
C &= \{F(\bm{x} + \text{ADB} \cdot \bm{d_1}) = y(\bm{x})\} = \{g(\bm{d_1}) > \text{ADB}\}\\
D &= \{F(\bm{x} + \text{ADB} \cdot \bm{d_2}) = y(\bm{x})\} = \{g(\bm{d_2}) > \text{ADB}\}
\end{cases}
\tag{4}
\end{align*}
We assume that \(g(\bm{d_1})\) and \(g(\bm{d_2})\) are independent random variables that follow an identical probability distribution \(\rho\). The eligibility of adopting this assumption is clarified in Appendix~\ref{appendix Same_distribution_function_rho}. The probability \(P(A),P(B),P(C),P(D)\) for any \(\text{ADB} \in [start, end]\) can be calculated respectively by:
\begin{equation}
\label{equation_PAPB_ab}
\begin{cases}
P(A) = P(B) = a = \int_{start}^{\text{ADB}} \rho(r) \, dr \\
P(C) = P(D) = b = \int_{\text{ADB}}^{end} \rho(r) \, dr
\end{cases}
\tag{5}
\end{equation}
Accordingly, \(P(B \cap C) = P(B) \cdot P(C \mid B)\) gives the probability of \(g(\bm{d_2}) \leq \text{ADB} < g(\bm{d_1})\), as illustrated in Figure~\ref{Fig1.sub.3}. To avoid the complexity of determining \( P(C \mid B) \), which requires extra queries, \(g(\bm{d_1})\) and \(g(\bm{d_2})\) are assumed to be independent. Therefore, 
\begin{equation}
\label{equation_assume_independent}
\begin{cases}
P(B \cap C) = P(B) \cdot P(C \mid B) = P(B) \cdot P(C) = ab \\
P(A \cap D) = P(A) \cdot P(D \mid A) = P(A) \cdot P(D) = ab\\
P(A \cap B) = P(A) \cdot P(B \mid A) = P(A) \cdot P(B) = a^2 \\
P(C \cap D) = P(C) \cdot P(D \mid C) = P(C) \cdot P(D) = b^2\\
\end{cases}
\tag{6}
\end{equation}
Consequently, in each comparison attempt, the probability \(P({SUCC})\) for ADB to successfully differentiate a pair of directions \(\bm{d_1}\) and \(\bm{d_2}\) and the probability \(P({FAIL})\) for ADB to fail to do so are determined respectively by:
\begin{equation}
\label{equation_Psucc_Pfail}
\begin{cases}
P({SUCC}) = P(B \cap C) + P(A \cap D) = 2ab, \\
P({FAIL}) = P(A \cap B) + P(C \cap D) = a^2 + b^2
\end{cases}
\tag{7}
\end{equation}
\begin{equation*}
    \text{with } P({SUCC}) + P({FAIL}) = a^2 + b^2 + 2ab = 1 \text{\ and\ } a+b=1
\end{equation*}
\(P({SUCC})\) reaches its maximum by setting \(a = b = 0.5\), resulting in \(P({SUCC}) = P({FAIL}) = 1/2\). Hence, according to Eqn. (\ref{equation_PAPB_ab}), ADB should be set to divide the probability density function \(\rho(r)\) into two parts of equal size, instead of setting ADB to \((start + end)/2\) at lines 6 and 9 in Algorithm~\ref{algorithm_compareADB}. Specifically, we can set ADB to satisfy the following equation, without querying the target model:
\begin{equation}
\label{equation_ADB_is_median_rho}
    \int_{start}^{\text{ADB}} \rho(r) \, dr = \int_{\text{ADB}}^{end} \rho(r) \, dr = \frac{1}{2} \cdot \int_{start}^{end} \rho(r) \, dr
    \tag{8}
\end{equation}
Especially, if the probability distribution is uniformly identical, then ADB should be set to the mid point between \(start\) and \(end\). The expected number of queries required for differentiating any pair of perturbation directions can be kept small. In fact, it is straightforward to show that \(P({SUCC})=1/2\). If differentiation fails in one comparison attempt, in the next attempt, Algorithm~\ref{algorithm_compareADB} updates \(start\) and \(end\) according to lines 7 and 10, and then sets ADB to satisfy Eqn. (\ref{equation_ADB_is_median_rho}). Hence, \(P(SUCC))\) remains to be \(1/2\). Assuming that differentiation succeeds after \( C \) comparison attempts, which means that the first \( C-1 \) comparison attempts fail and the \( C \)-th comparison attempt succeeds, the probability that \( C \) equals any given \( n \) is: \(
P(C = n) = (P({FAIL}))^{n-1} \cdot P({SUCC}) = a^{n-1} \cdot b = \left(\frac{1}{2}\right)^n\). Hence, the expected number of comparison attempts \(\bar{C}\) can be determined as \(\bar{C} = 1\cdot\left(\frac{1}{2}\right)^1 + 2\cdot\left(\frac{1}{2}\right)^2 \cdot\cdot\cdot + 
 n\cdot\left(\frac{1}{2}\right)^n = \sum_{n=1}^{\infty} n \cdot \left(\frac{1}{2}\right)^n = 2\). Therefore, 2 consecutive guesses of ADB on average are required to differentiate a pair of directions. Meanwhile, for each comparison attempt, a total of 2 queries are performed using the perturbations ADB\(\cdot \bm{d_1}\) and ADB\( \cdot \bm{d_2}\) respectively in line 1, 5, 8, 11, and 13 of Algorithm~\ref{algorithm_compareADB}. In total, the average number of queries required to differentiate any two directions is \(\bar{Q} = 2 \times \bar{C} = 2 \times 2 = 4\).

Since Eqn. (\ref{equation_assume_independent}) is obtained by assuming that \(g(\bm{d_1})\) and \(g(\bm{d_2})\) are independent, the actual number of queries required to differentiate any two directions may not be \(4\). However, the experiment results in Subsection~\ref{section experiment_result} show that the number of queries achieved by ADBA is very close to \(4\) in practice. Compared with RayS \cite{chen_rays_2020} that requires on average \(10\) queries to reach an accuracy level of \(0.001\approx 2^\text{-10}\) through binary search in our experiments, ADBA can noticeably reduce the required number of queries by approximately 2.5 times. 

In Appendix~\ref{appendix Same_distribution_function_rho} , in order to estimate the probability density function \(\rho\), we conduct a statistical analysis of the distribution of the actual decision boundaries \(r_{\text{best}}\), \(g(\bm{d_1})\) and \(g(\bm{d_2})\) under various settings of \(\bm{d_{\text{best}}}\), \(\bm{d_1}\) and \(\bm{d_2}\). Subsequently, we utilize an inverse proportional function, i.e., \(\rho(r) = \frac{a}{(|-r/r_{\text{best}} + d|)^b + c}\), \(r \in [0, r_{\text{best}}]\), to estimate the true distribution of the decision boundary, where \(a\), \(b\), \(c\), and \(d\) are the estimation parameters. Specifically, \(r_{\text{best}}\) is the ADB of the current best direction \(\bm{d_{\text{best}}}\), which serves as the input of Algorithm~\ref{algorithm_compareADB}. The efficiency of ADBA-md is insensitive to the accuracy of \(\rho(r)\). Hence, ADBA-md can be easily applied to other datasets and target models, as supported by more experiment results reported in Appendix~\ref{appendix ADBA-md mnist cifar}. ADBA and ADBA-md do not incorporate random factors in Algorithm \ref{algorithm_ADBA} and \ref{algorithm_compareADB}. Therefore, for the same original image, ADBA and ADBA-md will consistently generate the same adversarial examples across multiple repeated experiments.

\vspace{-0.2cm}
\section{Experiments}
\vspace{-0.2cm}
\label{section experiment}

\subsection{Experiment Settings}
\label{section experiment_settings}
In this section, we conduct comprehensive experiments to compare ADBA and ADBA-md against several state-of-the-art decision-based attack approaches, including OPT \cite{cheng_query-efficient_2018}, SignOPT \cite{cheng_sign-opt_2020}, HSJA \cite{chen_hopskipjumpattack_2020}, and RayS \cite{chen_rays_2020}. Our experiments are conducted on a server with an Intel Xeon Gold 6330 CPU, NVIDIA 4090 GPUs using PyTorch 2.3.0, Torchvision 0.18.0 on the Python 3.11.5 platform.

{\bf Datasets and target models}. We evaluate the performance of ADBA and ADBA-md on the ImageNet dataset~\cite{deng_imagenet_2009}, where each image in this dataset is resized to \(224 \times 224 \times 3\) as a common practice \cite{chen_rays_2020}. The dataset is under BSD 3-Clause License. Our evaluation includes six contemporary machine learning models that are commonly used as attack target models \cite{chen_rays_2020,cheng_sign-opt_2020,li_aha_2021,reza_cgba_2023}: VGG19 \cite{simonyan_very_2015}, ResNet50 \cite{he_deep_2016}, Inception-V3 \cite{szegedy_rethinking_2016}, ViT-B32 \cite{dosovitskiy_image_2021}, DenseNet161 \cite{huang_densely_2017}, and EfficientNet-B0 \cite{tan_efficientnet_2019}. These models are implemented through the \text{torchvision.models} package in Python. We adopted the latest model weights retrievable from \url{https:// download.pytorch.org/models/}: \text{vgg19-dcbb9e9d.pth}, \text{resnet50-11ad3fa6.pth}, \text{inception\_v3\_google-0cc3c7bd.pth}, \text{vit\_b\_32-d86f8d99.pth}, \text{efficientnet\_b0\_rwightman-7f5810bc.pth}, and \text{densenet161-8d451a50.pth}.

{\bf Attack settings}. Following cutting-edge research on black-box adversarial attacks \cite{chen_rays_2020,moon_parsimonious_2019}, we adopt the \(l_{\infty}\) norm and set the perturbation strength threshold \(\epsilon=0.05\). Meanwhile, the query budget is set to 10000 for each model \cite{chen_rays_2020}. The parameters in the distribution function \(\rho(r)=\frac{a}{(|-r/r_{\text{best}}+d|)^b+c}\), \(r\in[0,r_{\text{best}}]\) are set to be \(a = 0.0313\), \(b = 3.066\), \(c = 0.168\), and \(d = 1.134\) according to Appendix~\ref{appendix Same_distribution_function_rho}. The search tolerance threshold \(\tau=0.002\).

\vspace{-0.2cm}
\begin{table}[tb]
\centering
\caption{Comparison of decision-based untargeted attacks on the ImageNet dataset with 10,000 query budgets and maximum \(l_{\infty}\) perturbation strengths \(\epsilon=0.05\)}
\label{table:hard_label_attacks1}
\setlength{\tabcolsep}{3.3mm}
\begin{tabular}{l|cccccc}
\hline
\hspace{2em}\textbf{Model} & \textbf{VGG}& \textbf{ResNet}& \textbf{Inception}& \textbf{ViT}& \textbf{DenseNet}& \textbf{EfficientNet}\\ 
 \textbf{Attack   }& \textbf{19}& \textbf{50}& \textbf{-V3}& \textbf{-B32}& \textbf{161}&\textbf{-B0}\\ \hline
\textbf{OPT} & 1473.0& 2096.8& 2213.6& 3291.7& 3373.6& 2825.4\\
 & (1191)& (1387)& (1419)& (2158)& (2350)&(1813)\\ 
 & 25.1\% & 9.2\% & 18.1\% & 16.2\% & 12.9\% & 10.8\% \\ 
\textbf{SignOPT} & 2575.2& 3042.3& 2253.3& 3499.4& 3943.7& 3031.9\\
 & (1765)& (1727)& (1337)& (1969)& (3051)&(1897)\\
 & 42.5\% & 22.5\% & 35.0\% & 30.8\% & 25.2\% & 17.6\% \\ 
\textbf{HSJA} & 589.1& 1220.0& 733.6& 996.9& 1891.0& 1728.3\\
 & (324)& (401)& (358)& (422)& (812)&(559)\\
 & 29.5\% & 12.7\% & 21.9\% & 18.8\% & 16.1\% & 13.9\% \\ 
\textbf{RayS} & 470.1& 1243.5& 672.2& 1005.1& 733.6& 669.7\\
 & (295)& (721)& (339)& (414)& (358)&(321)\\
 & \textbf{100\%}& 98.8\% & 98.8\% & \textbf{98.8\%}& 99.1\% & \textbf{99.8\%}\\ 
\textbf{ADBA} & 237.3& 769.3& 461.3& 768.4& 485.7& 428.3\\
 & (119)& (339)& (156)& (221)& (158)&\textbf{(143)}\\
 & \textbf{100\%}& 97.9\% & 99.4\% & \textbf{98.8\%}& 99.2\% & 99.6\% \\ 
\textbf{ADBA-md} & \textbf{197.6}& \textbf{685.6}& \textbf{418.3}& \textbf{651.9}& \textbf{457.8}& \textbf{343.2}\\
 & \textbf{(115)}& \textbf{(273)}& \textbf{(149)}& \textbf{(202)}& \textbf{(149)}&(146)\\
 & \textbf{100\%} & \textbf{99.3\%}& \textbf{99.7\%}& \textbf{99.8\%}& \textbf{99.7\%}& \textbf{99.8\%}\\ \hline
\end{tabular}
\end{table}

\begin{figure}[tb]
\centering
\includegraphics[width=1.0\textwidth]{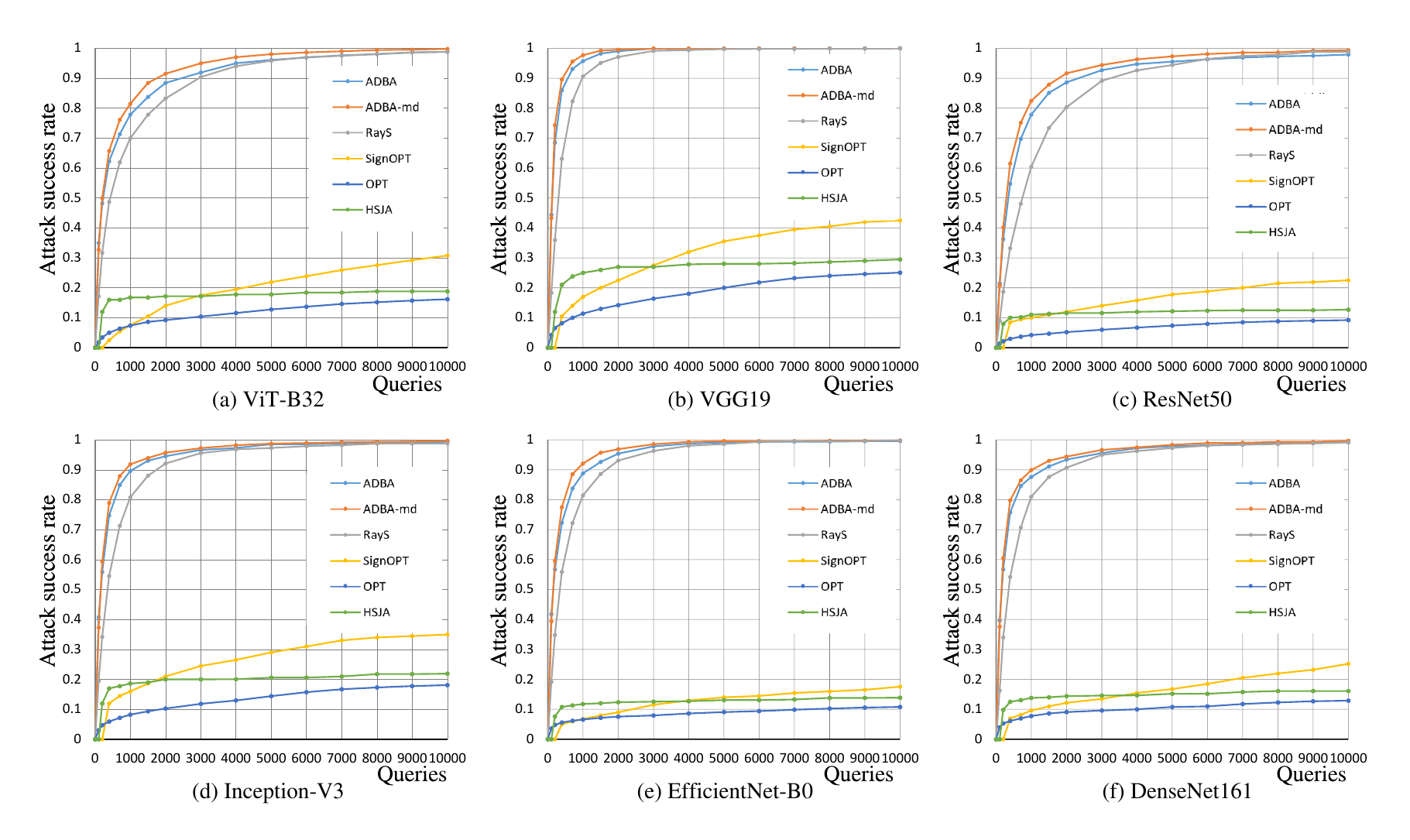}
\caption{Attack success rates versus number of queries for six hard-label attacks on ImageNet.}
\label{fig:compare_acc_line}
\end{figure}

\vspace{-0.2cm}
\subsection{Experiment Results}
\label{section experiment_result}

For six pre-trained models, we randomly select 1,000 correctly classified images from the ImageNet test set for each model. Then we produce adversarial examples for each image by using six hard-label attacks respectively to determine the attack success rate, which is calculated by \(\frac{|E_{\text{successful}}|}{|E|} \). Here, \(E\) refers to the set of all test cases (\(|E|=1000\)) and \(E_{\text{successful}}\) is the set of successful attacks that meet the requirements for the query budget (i.e., 10000) and the perturbation strength threshold (i.e., \(\epsilon=0.05\)). In Table~\ref{table:hard_label_attacks1}, we summarize the attack success rate and the average (median) number of queries for six attack methods across six target models. ADBA-md provides the highest attack success rate, exceeding 99.3\% across all target models, and requires the fewest average number of queries across six black-box attacks. ADBA method results in the smallest median number of queries on the EfficientNet-B0 model, and ADBA-md has the smallest median number of queries on other models. Compared to other attack strategies, our methods significantly reduce the average number of queries to below 800 and the median number of queries to below 400.

Under similar attack success rates, both ADBA and ADBA-md can significantly reduce the average and median number of queries compared to RayS. Specifically, the average number of queries is approximately 60-70\% of those required by RayS, and the median is reduced by 50\%. Our results indicate that it is unnecessary to perform precise binary search as in RayS. Instead, it is much more efficient to optimize perturbation directions by using approximation decision boundaries.

Figure~\ref{fig:compare_acc_line} presents a comparative performance analysis of six black-box methods on six target models, focusing on the attack success rate relative to the number of queries. ADBA and ADBA-md consistently achieve the highest attack success rates compared to other approaches. Notably, even with a very low query budget (1000 queries), both ADBA and ADBA-md attain an attack success rate of approximately 90\%, whereas RayS achieves around 80\%. The remaining three attack methods fail to obtain an attack success rate above 30\%. This result confirms that our methods can achieve substantially higher success rate, making them highly effective under restricted query budgets.

\begin{table}[tb]
    \centering
    \caption{The Average number of iterations and queries performed by ADBA and ADBA-md on six contemporary image classification models.}
    \begin{tabular}{l|ccccccc}
        \hline
          & & \textbf{VGG}& \textbf{ResNet}& \textbf{Inception}& \textbf{ViT}& \textbf{DenseNet}&\textbf{EfficientNet}\\
 & & \textbf{19}& \textbf{50}& \textbf{-V3}& \textbf{-B32}&\textbf{161}&\textbf{-B0}\\
        \hline
        \textbf{ADBA}& \textbf{Iterations}& 49.90 & 168.98 & 103.25 & 170.42 & 112.87  &94.34\\
        & \textbf{Q each I}& 4.755 & 4.552 & 4.468 & 4.509 & 4.303  &4.540\\
        & \textbf{Queries}& 237.3 & 769.3 & 461.3 & 768.4 & 485.7  &428.3\\
        \hline
        \textbf{ADBA}& \textbf{Iterations}& 50.92 & 197.46 & 118.22 & 184.31 & 133.22  &95.39\\
        \textbf{-md}& \textbf{\% Change}& \textup{↑2.0\%} & \textup{↑16.9\%} & \textup{↑14.5\%} & \textup{↑8.2\%} & \textup{↑18.0\%}  &\textup{↑1.1\%}\\
        & \textbf{Q each I}& 3.880 & 3.472 & 3.538 & 3.537 & 3.436  &3.598\\
        & \textbf{\% Change}& \textup{↓18.4\%} & \textup{↓23.7\%} & \textup{↓20.8\%} & \textup{↓21.6\%} & \textup{↓20.1\%}  &\textup{↓20.7\%}\\
        & \textbf{Queries}& 197.6 & 685.6 & 418.3 & 651.9 & 457.8  &343.2\\
        & \textbf{\% Change}& \textup{↓16.7\%} & \textup{↓10.9\%} & \textup{↓9.32\%} & \textup{↓15.2\%} & \textup{↓5.74\%}  &\textup{↓19.9\%}\\
        \hline
    \end{tabular}
\label{table-2}
\end{table}

In Table~\ref{table-2}, we analyze the impact of the median search method (see Subsection~\ref{section optimize ADB}) on the efficiency of ADBA-md. For all six models, ADBA-md requires a higher average number of search iterations (indicated by the upward arrows in \%Change of Iteration), yet it has less number of queries per iteration (shown with downward arrows in \%Change of Q each I). Consequently, ADBA-md requires less number queries in total (as denoted by the downward arrows in \%Change of Queries ). The increase in iterations for ADBA-md is attributed to the reduced number of queries in each iteration, leading to a smaller improvement in the ADBs per iteration. Therefore, more iterations are required to reach the perturbation strength threshold \(\epsilon\). Approaches proposed in this paper are only used for the study of adversarial machine learning and the robustness of machine learning models, and does not target any real system. There is no potential negative impact.

\vspace{-0.2cm}
\section{Conclusions}
\vspace{-0.2cm}
\label{section conclusion}

Decision-based black-box attacks are particularly relevant in practice as they rely solely on the hard label to generate adversarial examples. In this paper, we proposed a new Approximation Decision Boundary Approach (ADBA) for query-efficient decision-based attacks. Our approach introduced a novel insight: it is feasible to compare and optimize perturbation directions even without exact knowledge of the decision boundaries. Specifically, given any Approximation Decision Boundary (ADB), if one perturbation direction causes the model to fail while the other direction does not, the former direction is deemed superior. Driven by this insight that has been consistently overlooked in the past research, we substantially improved the direction search framework by replacing the binary search for precise decision boundaries with a query-efficient method that searches for ADBs. Subsequently, we analyzed the distribution of decision boundaries and developed ADBA-md to noticeably improve the efficiency of approximating decision boundaries. Our experiments on six image classification models clearly validated the effectiveness of our ADBA and ADBA-md approaches in enhancing the attack success rate while limiting the number of queries.

\printbibliography

@article{ghosh_black-box_2022,
	title = {A black-box adversarial attack strategy with adjustable sparsity and generalizability for deep image classifiers},
	volume = {122},
	journal = {Pattern Recognition},
	author = {Ghosh, Arka and Mullick, Sankha Subhra and Datta, Shounak},
	year = {2022},
	note = {Publisher: Elsevier},
	pages = {108279}
}

@article{shi_decision-based_2022,
	title = {Decision-based black-box attack against vision transformers via patch-wise adversarial removal},
	volume = {35},
	journal = {Advances in Neural Information Processing Systems},
	author = {Shi, Yucheng and Han, Yahong and Tan, Yu-an and Kuang, Xiaohui},
	year = {2022},
	pages = {12921--12933}
}

@incollection{avidan_triangle_2022,
	address = {Cham},
	title = {Triangle {Attack}: {A} {Query}-{Efficient} {Decision}-{Based} {Adversarial} {Attack}},
	volume = {13665},
	shorttitle = {Triangle {Attack}},
	language = {en},
	booktitle = {Computer {Vision} – {ECCV} 2022},
	publisher = {Springer Nature Switzerland},
	editor = {Avidan, Shai and Brostow, Gabriel and Cissé, Moustapha},
        author={Wang, Xiaosen and Zhang, Zeliang and Tong, Kangheng and Gong, Dihong and He, Kun and Li, Zhifeng and Liu, Wei},
	year = {2022},
	note = {Series Title: Lecture Notes in Computer Science},
	pages = {156--174}
}

@misc{brendel_decision-based_2018,
	title = {Decision-{Based} {Adversarial} {Attacks}: {Reliable} {Attacks} {Against} {Black}-{Box} {Machine} {Learning} {Models}},
	shorttitle = {Decision-{Based} {Adversarial} {Attacks}},
	publisher = {arXiv},
	author = {Brendel, Wieland and Rauber, Jonas and Bethge, Matthias},
	month = feb,
	year = {2018},
	note = {arXiv:1712.04248 [cs, stat]}
}

@article{lin_sensitive_2023,
	title = {Sensitive region-aware black-box adversarial attacks},
	volume = {637},
	issn = {00200255},
	language = {en},
	journal = {Information Sciences},
	author = {Lin, Chenhao and Han, Sicong and Zhu, Jiongli and Li, Qian and Shen, Chao and Zhang, Youwei and Guan, Xiaohong},
	month = aug,
	year = {2023},
	pages = {118929}
}

@inproceedings{brunner_guessing_2019,
	title = {Guessing smart: {Biased} sampling for efficient black-box adversarial attacks},
	shorttitle = {Guessing smart},
	booktitle = {Proceedings of the {IEEE}/{CVF} {International} {Conference} on {Computer} {Vision}},
	author = {Brunner, Thomas and Diehl, Frederik and Le, Michael Truong and Knoll, Alois},
	year = {2019},
	pages = {4958--4966}
}

@inproceedings{feng_boosting_2022,
	title = {Boosting black-box attack with partially transferred conditional adversarial distribution},
	booktitle = {Proceedings of the {IEEE}/{CVF} {Conference} on {Computer} {Vision} and {Pattern} {Recognition}},
	author = {Feng, Yan and Wu, Baoyuan and Fan, Yanbo and Liu, Li and Li, Zhifeng and Xia, Shu-Tao},
	year = {2022},
	pages = {15095--15104}
}

@inproceedings{chen_hopskipjumpattack_2020,
	title = {Hopskipjumpattack: {A} query-efficient decision-based attack},
	shorttitle = {Hopskipjumpattack},
	booktitle = {2020 ieee symposium on security and privacy (sp)},
	publisher = {IEEE},
	author = {Chen, Jianbo and Jordan, Michael I. and Wainwright, Martin J.},
	year = {2020},
	pages = {1277--1294},
}

@article{bai_query_2023,
	title = {Query efficient black-box adversarial attack on deep neural networks},
	volume = {133},
	journal = {Pattern Recognition},
	author = {Bai, Yang and Wang, Yisen and Zeng, Yuyuan},
	year = {2023},
	note = {Publisher: Elsevier},
	pages = {109037}
}

@inproceedings{rahmati_geoda_2020,
	title = {Geoda: a geometric framework for black-box adversarial attacks},
	shorttitle = {Geoda},
	booktitle = {Proceedings of the {IEEE}/{CVF} conference on computer vision and pattern recognition},
	author = {Rahmati, Ali and Moosavi-Dezfooli, Seyed-Mohsen and Frossard, Pascal and Dai, Huaiyu},
	year = {2020},
	pages = {8446--8455}
}

@inproceedings{chen_rays_2020,
	address = {Virtual Event CA USA},
	title = {{RayS}: {A} {Ray} {Searching} {Method} for {Hard}-label {Adversarial} {Attack}},
	isbn = {978-1-4503-7998-4},
	shorttitle = {{RayS}},
	language = {en},
	booktitle = {Proceedings of the 26th {ACM} {SIGKDD} {International} {Conference} on {Knowledge} {Discovery} \& {Data} {Mining}},
	publisher = {ACM},
	author = {Chen, Jinghui and Gu, Quanquan},
	month = aug,
	year = {2020},
	pages = {1739--1747}
}

@inproceedings{reza_cgba_2023,
	title = {{CGBA}: {Curvature}-aware geometric black-box attack},
	shorttitle = {{CGBA}},
	booktitle = {Proceedings of the {IEEE}/{CVF} {International} {Conference} on {Computer} {Vision}},
	author = {Reza, Md Farhamdur and Rahmati, Ali and Wu, Tianfu and Dai, Huaiyu},
	year = {2023},
	pages = {124--133}
}

@misc{cheng_sign-opt_2020,
	title = {Sign-{OPT}: {A} {Query}-{Efficient} {Hard}-label {Adversarial} {Attack}},
	shorttitle = {Sign-{OPT}},
	publisher = {arXiv},
	author = {Cheng, Minhao and Singh, Simranjit and Chen, Patrick and Chen, Pin-Yu and Liu, Sijia and Hsieh, Cho-Jui},
	month = feb,
	year = {2020},
	note = {arXiv:1909.10773 [cs, stat]}
}

@inproceedings{liu_geometry-inspired_2019,
	title = {A geometry-inspired decision-based attack},
	booktitle = {Proceedings of the {IEEE}/{CVF} {International} {Conference} on {Computer} {Vision}},
	author = {Liu, Yujia and Moosavi-Dezfooli, Seyed-Mohsen and Frossard, Pascal},
	year = {2019},
	pages = {4890--4898}
}

@inproceedings{li_qeba_2020,
	title = {Qeba: {Query}-efficient boundary-based blackbox attack},
	shorttitle = {Qeba},
	booktitle = {Proceedings of the {IEEE}/{CVF} conference on computer vision and pattern recognition},
	author = {Li, Huichen and Xu, Xiaojun and Zhang, Xiaolu and Yang, Shuang and Li, Bo},
	year = {2020},
	pages = {1221--1230}
}

@inproceedings{maho_surfree_2021,
	title = {Surfree: a fast surrogate-free black-box attack},
	shorttitle = {Surfree},
	booktitle = {Proceedings of the {IEEE}/{CVF} {Conference} on {Computer} {Vision} and {Pattern} {Recognition}},
	author = {Maho, Thibault and Furon, Teddy and Le Merrer, Erwan},
	year = {2021},
	pages = {10430--10439}
}

@inproceedings{li_aha_2021,
	title = {Aha! adaptive history-driven attack for decision-based black-box models},
	booktitle = {Proceedings of the {IEEE}/{CVF} {International} {Conference} on {Computer} {Vision}},
	author = {Li, Jie and Ji, Rongrong and Chen, Peixian and Zhang, Baochang and Hong, Xiaopeng and Zhang, Ruixin and Li, Shaoxin and Li, Jilin and Huang, Feiyue and Wu, Yongjian},
	year = {2021},
	pages = {16168--16177}
}

@inproceedings{eykholt_robust_2018,
	title = {Robust physical-world attacks on deep learning visual classification},
	booktitle = {Proceedings of the {IEEE} conference on computer vision and pattern recognition},
	author = {Eykholt, Kevin and Evtimov, Ivan and Fernandes, Earlence and Li, Bo and Rahmati, Amir and Xiao, Chaowei and Prakash, Atul and Kohno, Tadayoshi and Song, Dawn},
	year = {2018},
	pages = {1625--1634}
}

@inproceedings{li_towards_2023,
	title = {Towards benchmarking and assessing visual naturalness of physical world adversarial attacks},
	booktitle = {Proceedings of the {IEEE}/{CVF} {Conference} on {Computer} {Vision} and {Pattern} {Recognition}},
	author = {Li, Simin and Zhang, Shuning and Chen, Gujun and Wang, Dong and Feng, Pu and Wang, Jiakai and Liu, Aishan and Yi, Xin and Liu, Xianglong},
	year = {2023},
	pages = {12324--12333}
}

@inproceedings{li_physical-world_2023,
	title = {Physical-world optical adversarial attacks on 3d face recognition},
	booktitle = {Proceedings of the {IEEE}/{CVF} {Conference} on {Computer} {Vision} and {Pattern} {Recognition}},
	author = {Li, Yanjie and Li, Yiquan and Dai, Xuelong and Guo, Songtao and Xiao, Bin},
	year = {2023},
	pages = {24699--24708}
}

@inproceedings{xie_feature_2019,
	title = {Feature denoising for improving adversarial robustness},
	booktitle = {Proceedings of the {IEEE}/{CVF} conference on computer vision and pattern recognition},
	author = {Xie, Cihang and Wu, Yuxin and Maaten, Laurens van der and Yuille, Alan L. and He, Kaiming},
	year = {2019},
	pages = {501--509}
}

@inproceedings{xie_adversarial_2017,
	title = {Adversarial examples for semantic segmentation and object detection},
	booktitle = {Proceedings of the {IEEE} international conference on computer vision},
	author = {Xie, Cihang and Wang, Jianyu and Zhang, Zhishuai and Zhou, Yuyin and Xie, Lingxi and Yuille, Alan},
	year = {2017},
	pages = {1369--1378}
}

@article{long_survey_2022,
	title = {A survey on adversarial attacks in computer vision: {Taxonomy}, visualization and future directions},
	volume = {121},
	issn = {0167-4048},
	shorttitle = {A survey on adversarial attacks in computer vision},
	journal = {Computers \& Security},
	author = {Long, Teng and Gao, Qi and Xu, Lili and Zhou, Zhangbing},
	month = oct,
	year = {2022},
	pages = {102847}
}

@inproceedings{wang_enhancing_2021,
	title = {Enhancing the {Transferability} of {Adversarial} {Attacks} {Through} {Variance} {Tuning}},
	language = {en},
	author = {Wang, Xiaosen and He, Kun},
	year = {2021},
	pages = {1924--1933}
}

@inproceedings{chen_zoo_2017,
	address = {Dallas Texas USA},
	title = {{ZOO}: {Zeroth} {Order} {Optimization} {Based} {Black}-box {Attacks} to {Deep} {Neural} {Networks} without {Training} {Substitute} {Models}},
	isbn = {978-1-4503-5202-4},
	shorttitle = {{ZOO}},
	language = {en},
	booktitle = {Proceedings of the 10th {ACM} {Workshop} on {Artificial} {Intelligence} and {Security}},
	publisher = {ACM},
	author = {Chen, Pin-Yu and Zhang, Huan and Sharma, Yash and Yi, Jinfeng and Hsieh, Cho-Jui},
	month = nov,
	year = {2017},
	pages = {15--26}
}

@inproceedings{cheng_query-efficient_2018,
	title = {Query-{Efficient} {Hard}-label {Black}-box {Attack}: {An} {Optimization}-based {Approach}},
	shorttitle = {Query-{Efficient} {Hard}-label {Black}-box {Attack}},
	booktitle = {International {Conference} on {Learning} {Representations}},
	author = {Cheng, Minhao and Le, Thong and Chen, Pin-Yu and Zhang, Huan and Yi, JinFeng and Hsieh, Cho-Jui},
	year = {2018}
}

@inproceedings{ilyas_black-box_2018,
	title = {Black-box adversarial attacks with limited queries and information},
	booktitle = {International conference on machine learning},
	publisher = {PMLR},
	author = {Ilyas, Andrew and Engstrom, Logan and Athalye, Anish and Lin, Jessy},
	year = {2018},
	pages = {2137--2146}
}

@inproceedings{chen_frank-wolfe_2020,
	title = {A frank-wolfe framework for efficient and effective adversarial attacks},
	volume = {34},
	booktitle = {Proceedings of the {AAAI} conference on artificial intelligence},
	author = {Chen, Jinghui and Zhou, Dongruo and Yi, Jinfeng and Gu, Quanquan},
	year = {2020},
	note = {Issue: 04},
	pages = {3486--3494}
}

@inproceedings{moon_parsimonious_2019,
	title = {Parsimonious black-box adversarial attacks via efficient combinatorial optimization},
	booktitle = {International conference on machine learning},
	publisher = {PMLR},
	author = {Moon, Seungyong and An, Gaon and Song, Hyun Oh},
	year = {2019},
	pages = {4636--4645}
}

@inproceedings{deng_imagenet_2009,
	title = {{ImageNet}: {A} large-scale hierarchical image database},
	shorttitle = {{ImageNet}},
	booktitle = {2009 {IEEE} {Conference} on {Computer} {Vision} and {Pattern} {Recognition}},
	author = {Deng, Jia and Dong, Wei and Socher, Richard and Li, Li-Jia and Li, Kai and Fei-Fei, Li},
	month = jun,
	year = {2009},
	note = {ISSN: 1063-6919},
	pages = {248--255},
}

@article{simonyan_very_2015,
	title = {Very deep convolutional networks for large-scale image recognition},
	language = {en},
	journal = {3rd International Conference on Learning Representations (ICLR 2015)},
	author = {Simonyan, K. and Zisserman, A.},
	year = {2015},
	note = {Publisher: Computational and Biological Learning Society}
}

@inproceedings{he_deep_2016,
	title = {Deep residual learning for image recognition},
	booktitle = {Proceedings of the {IEEE} conference on computer vision and pattern recognition},
	author = {He, Kaiming and Zhang, Xiangyu and Ren, Shaoqing and Sun, Jian},
	year = {2016},
	pages = {770--778}
}

@inproceedings{szegedy_rethinking_2016,
	title = {Rethinking the inception architecture for computer vision},
	booktitle = {Proceedings of the {IEEE} conference on computer vision and pattern recognition},
	author = {Szegedy, Christian and Vanhoucke, Vincent and Ioffe, Sergey and Shlens, Jon and Wojna, Zbigniew},
	year = {2016},
	pages = {2818--2826}
}

@misc{dosovitskiy_image_2021,
	title = {An {Image} is {Worth} 16x16 {Words}: {Transformers} for {Image} {Recognition} at {Scale}},
	shorttitle = {An {Image} is {Worth} 16x16 {Words}},
	publisher = {arXiv},
	author = {Dosovitskiy, Alexey and Beyer, Lucas and Kolesnikov, Alexander and Weissenborn, Dirk and Zhai, Xiaohua and Unterthiner, Thomas and Dehghani, Mostafa and Minderer, Matthias and Heigold, Georg and Gelly, Sylvain and Uszkoreit, Jakob and Houlsby, Neil},
	month = jun,
	year = {2021},
	note = {arXiv:2010.11929 [cs]}
}

@inproceedings{huang_densely_2017,
	title = {Densely connected convolutional networks},
	booktitle = {Proceedings of the {IEEE} conference on computer vision and pattern recognition},
	author = {Huang, Gao and Liu, Zhuang and Van Der Maaten, Laurens and Weinberger, Kilian Q.},
	year = {2017},
	pages = {4700--4708}
}

@inproceedings{tan_efficientnet_2019,
	title = {{EfficientNet}: {Rethinking} {Model} {Scaling} for {Convolutional} {Neural} {Networks}},
	shorttitle = {{EfficientNet}},
	language = {en},
	booktitle = {Proceedings of the 36th {International} {Conference} on {Machine} {Learning}},
	publisher = {PMLR},
	author = {Tan, Mingxing and Le, Quoc},
	month = may,
	year = {2019},
	note = {ISSN: 2640-3498},
	pages = {6105--6114}
}

@article{li2020deep,
  title={Deep facial expression recognition: A survey},
  author={Li, Shan and Deng, Weihong},
  journal={IEEE transactions on affective computing},
  volume={13},
  number={3},
  pages={1195--1215},
  year={2020},
  publisher={IEEE}
}

@article{lecun1998mnist,
  title={Gradient-based learning applied to document recognition},
  author={LeCun, Yann and Bottou, L{\'e}on and Bengio, Yoshua and Haffner, Patrick},
  journal={Proceedings of the IEEE},
  volume={86},
  number={11},
  pages={2278--2324},
  year={1998},
  publisher={IEEE}
}

@article{krizhevsky2009cifar,
  title={Learning multiple layers of features from tiny images},
  author={Krizhevsky, Alex and Hinton, Geoffrey and others},
  year={2009},
  publisher={Toronto, ON, Canada}
}

@inproceedings{sehwag2019analyzing,
  title={Analyzing the robustness of open-world machine learning},
  author={Sehwag, Vikash and Bhagoji, Arjun Nitin and Song, Liwei and Sitawarin, Chawin and Cullina, Daniel and Chiang, Mung and Mittal, Prateek},
  booktitle={Proceedings of the 12th ACM Workshop on Artificial Intelligence and Security},
  pages={105--116},
  year={2019}
}

@InProceedings{ImprovingAdversarialRobustness2019,
  title = 	 {Improving Adversarial Robustness via Promoting Ensemble Diversity},
  author =       {Pang, Tianyu and Xu, Kun and Du, Chao and Chen, Ning and Zhu, Jun},
  booktitle = 	 {Proceedings of the 36th International Conference on Machine Learning},
  pages = 	 {4970--4979},
  year = 	 {2019},
  editor = 	 {Chaudhuri, Kamalika and Salakhutdinov, Ruslan},
  volume = 	 {97},
  series = 	 {Proceedings of Machine Learning Research},
  publisher =    {PMLR}
}

@inproceedings {287188,
author = {Heng Li and Zhang Cheng and Bang Wu and Liheng Yuan and Cuiying Gao and Wei Yuan and Xiapu Luo},
title = {Black-box Adversarial Example Attack towards {FCG} Based Android Malware Detection under Incomplete Feature Information},
booktitle = {32nd USENIX Security Symposium (USENIX Security 23)},
year = {2023},
isbn = {978-1-939133-37-3},
address = {Anaheim, CA},
pages = {1181--1198},
publisher = {USENIX Association},
month = aug
}

@InProceedings{wang2023ICCV,
    author    = {Wang, Ningfei and Luo, Yunpeng and Sato, Takami and Xu, Kaidi and Chen, Qi Alfred},
    title     = {Does Physical Adversarial Example Really Matter to Autonomous Driving? Towards System-Level Effect of Adversarial Object Evasion Attack},
    booktitle = {Proceedings of the IEEE/CVF International Conference on Computer Vision (ICCV)},
    year      = {2023},
    pages     = {4412-4423}
}

@inproceedings{wang2022poster,
  title={Poster: On the system-level effectiveness of physical object-hiding adversarial attack in autonomous driving},
  author={Wang, Ningfei and Luo, Yunpeng and Sato, Takami and Xu, Kaidi and Chen, Qi Alfred},
  booktitle={Proceedings of the 2022 ACM SIGSAC Conference on Computer and Communications Security},
  pages={3479--3481},
  year={2022}
}


\appendix

\section{Potential societal impacts}
\label{appendix societal impacts}
Adversarial attacks in the realm of image classification pose significant challenges yet offer substantial benefits to society. For image classification tasks, tiny additive perturbations in the input images can significantly affect the classification accuracy of a pre-trained model, highlighting the inherent vulnerabilities of these systems \cite{eykholt_robust_2018,lin_sensitive_2023}. Black-box adversarial attack methods, including the query-efficient ADBA proposed in this paper, exploit these vulnerabilities to generate adversarial samples that cause the models to misclassify, potentially creating security risks in real-world applications of image classification models \cite{sehwag2019analyzing}.

However, these challenges also bring about important benefits, particularly in enhancing the robustness of machine learning models \cite{ImprovingAdversarialRobustness2019}. By deliberately introducing perturbations that exploit system vulnerabilities, researchers can identify and rectify these weaknesses, thereby fortifying the models against potential malicious attacks. This proactive approach to security is crucial in applications such as autonomous driving systems, where the accuracy and reliability of machine learning predictions are paramount. For instance, in autonomous vehicles, employing adversarial attacks to test and improve systems can significantly contribute to public safety by ensuring that the vehicles can correctly interpret and react to real-world visual data under diverse conditions \cite{287188,wang2023ICCV,wang2022poster}. Thus, the integration of adversarial attacks into the development lifecycle of machine learning models not only advances technological reliability but also plays a critical role in safeguarding societal welfare by preventing failures that could have dire consequences.

\section{Additional related work}
\label{appendix Related work}

\subsection{Black-box attacks}
Black-box attacks can be divided into \emph{transfer-based attacks}, \emph{score-based attacks}, and \emph{decision-based attacks}. Transfer-based attacks \cite{feng_boosting_2022,ghosh_black-box_2022} require to use the training data of the target model to train a substitute model, then produce adversarial examples on substitute model using white-box attacks. Score-based attacks \cite{chen_zoo_2017} do not need to train a substitute model. Instead, they query the target model and craft the adversarial perturbation using model’s output scores, such as class probabilities or pre-softmax logits. However, in practice, attackers may only have access to the hard label (the label with the highest probability) outputs from the target model, rather than the associated scores. To cope with this limitation, decision-based attacks \cite{brendel_decision-based_2018,li_aha_2021,shi_decision-based_2022,chen_hopskipjumpattack_2020,cheng_sign-opt_2020} rely solely on the hard label from the target model to create adversarial examples.

\subsection{Decision-based black-block attacks}
Decision-based attacks can be divided into random search attacks, gradient estimation attacks, and geometric modeling attacks. In Section~\ref{section Related_work}, we talk about the random search attacks which is most related to our work. The following is the overview of gradient estimation attacks and geometric modeling attacks.

{\bf Gradient estimation attacks}. Gradient estimation attacks depend on estimating the normal vector at boundary points to direct perturbation efforts, which requires a large number of queries. HSJA \cite{chen_hopskipjumpattack_2020} estimates gradient directions using binary decision boundary information, enabling efficient adversarial example searches with fewer queries. qFool \cite{liu_geometry-inspired_2019} observes that decision boundaries typically have minor curvature around adversarial examples, simplifying gradient estimation to efficiently create imperceptible attacks. GeoDA \cite{rahmati_geoda_2020} notes that DNN decision boundaries usually exhibit small mean curvatures near data samples, allowing for effective normal vector estimation with minimal queries. QEBA \cite{li_qeba_2020} introduces three novel subspace optimization methods to cut down the number of queries from spatial, frequency, and intrinsic components perspectives.

{\bf Geometric modeling attacks}. Attackers can utilize geometric modeling of the perturbation vectors to simultaneously optimize the perturbation direction and strength. To avoid the estimations of gradient, The SurFree method \cite{maho_surfree_2021} tests various directions to quickly identify optimal perturbation paths. Triangle Attack \cite{avidan_triangle_2022} uses geometric principles and the law of sines to enhance precision in iterative attacks, focusing on low-frequency spaces for effective dimensionality reduction. CGBA \cite{reza_cgba_2023} maneuvers along a semicircular path on a constrained 2D plane to find new boundary points, ensuring efficiency despite geometric complexities.

\begin{table}[htb]
\centering
\captionsetup{justification=raggedright, singlelinecheck=false}
\caption{Comparison of ADBA and ADBA-CCM}
\label{table-ESMvsCCM}
\setlength{\tabcolsep}{3.3mm}
\begin{tabular}{l|cccccc}
\hline
\hspace{2em}\textbf{Models}& \textbf{VGG}& \textbf{ResNet}& \textbf{Inception}& \textbf{ViT}& \textbf{DenseNet}& \textbf{EfficientNet}\\
 \textbf{Attacks   }& \textbf{19}& \textbf{50}& \textbf{-V3}& \textbf{-B32}& \textbf{161}&\textbf{-B0}\\  \hline
\textbf{ADBA}& \textbf{237.3}& \textbf{769.3}& \textbf{461.3}& \textbf{768.4}& \textbf{485.7}& \textbf{428.3}\\
 & \textbf{(119)}& \textbf{(339)}& \textbf{(156)}& \textbf{(221)}& \textbf{(158)}&\textbf{(143)}\\
 & \textbf{100\%}& \textbf{97.9\%}& \textbf{99.4\%}& \textbf{98.8\%}& \textbf{99.2\%}& \textbf{99.6\%}\\ 
\textbf{ADBA-CCM}& 252.9& 843.1& 511.7& 909.4& 535.2& 474.0\\
 & (131)& (387)& (183)& (251)& (179)&163\\
 & \textbf{100\%}& 97.1\%& 99.1\%& 98.2\%& 98.7\%& 99.4\%\\ \hline
\end{tabular}
\end{table}

\section{Comparison between ADBA and ADBA with Continue Comparison Method}
\label{appendix ADBA-CCM}
In Section~\ref{section optimize ADB}, in line with Figures~\ref{Fig1.sub.1} and~\ref{Fig1.sub.3}, we argued that it is unnecessary to compare the current best direction \(\bm{d_2}\) with the previous best direction \(\bm{d_\text{best}}\). To support this argument, we conduct additional experiments on the Continue Comparison Method, i.e., ADBA-CCM. Specifically, after comparing \(g(\bm{d_1})\) and \(g(\bm{d_2})\) in line 1-16 of Algorithm~\ref{algorithm_compareADB}, ADBA-CCM performs extra comparisons between  \(g(\bm{d_\text{best}})\) and \(\text{min}(g(\bm{d_1}), g(\bm{d_2}))\).

Under identical experiment settings in Subsection~\ref{section experiment_settings}, Table~\ref{table-ESMvsCCM} reports the attack success rate and the average (median) number of queries achieved by ADBA and ADBA-CCM respectively. The results indicate that ADBA-CCM needs more queries both on average and in median, and has a lower attack success rate compared to ADBA, demonstrating ADBA's higher efficiency. The reason ADBA-CCM performs poorly is because it requires significantly more queries to compare \( g(\bm{d_\text{best}}) \) and \(\text{min}(g(\bm{d_1}), g(\bm{d_2}))\) during the early stages of direction optimization iterations. However, the corresponding improvement in \( g(\bm{d_\text{best}}) \) is minimal, leading to low efficiency and wasted queries.

\section{Analysis of the Distribution Function \texorpdfstring{\(\rho(r)\)}{rho(r)}}
\label{appendix Same_distribution_function_rho}
In Subsection~\ref{section optimize ADB}, we assume that the decision boundaries of \(\bm{d_1}\) and \(\bm{d_2}\) follow the same distribution \(\rho(r)\). In this appendix, we provide empirical evidence to support this assumption.

In Algorithm~\ref{algorithm_compareADB}, we use ADB to compare \(\bm{d_\text{best}}\), \(\bm{d_1}\), and \(\bm{d_2}\) without precisely identifying \(r_\text{best}\), \(g(\bm{d_1})\), and \(g(\bm{d_2})\). In this appendix, we choose to use the binary search method developed in RayS \cite{chen_rays_2020} (instead of Algorithm~\ref{algorithm_compareADB}) to precisely determine \(r_\text{best}\), \(g(\bm{d_1})\), and \(g(\bm{d_2})\) while running Algorithm~\ref{algorithm_ADBA}. The target model \(F\) is ResNet50, and the dataset consists of 100 randomly selected images from ImageNet. We then analyze the distribution of \(g(\bm{d_1})\) and \(g(\bm{d_2})\) within the range \( [0, r_{\text{best}}]\).

\begin{figure}[ht]
    \centering
    \includegraphics[width=1\linewidth]{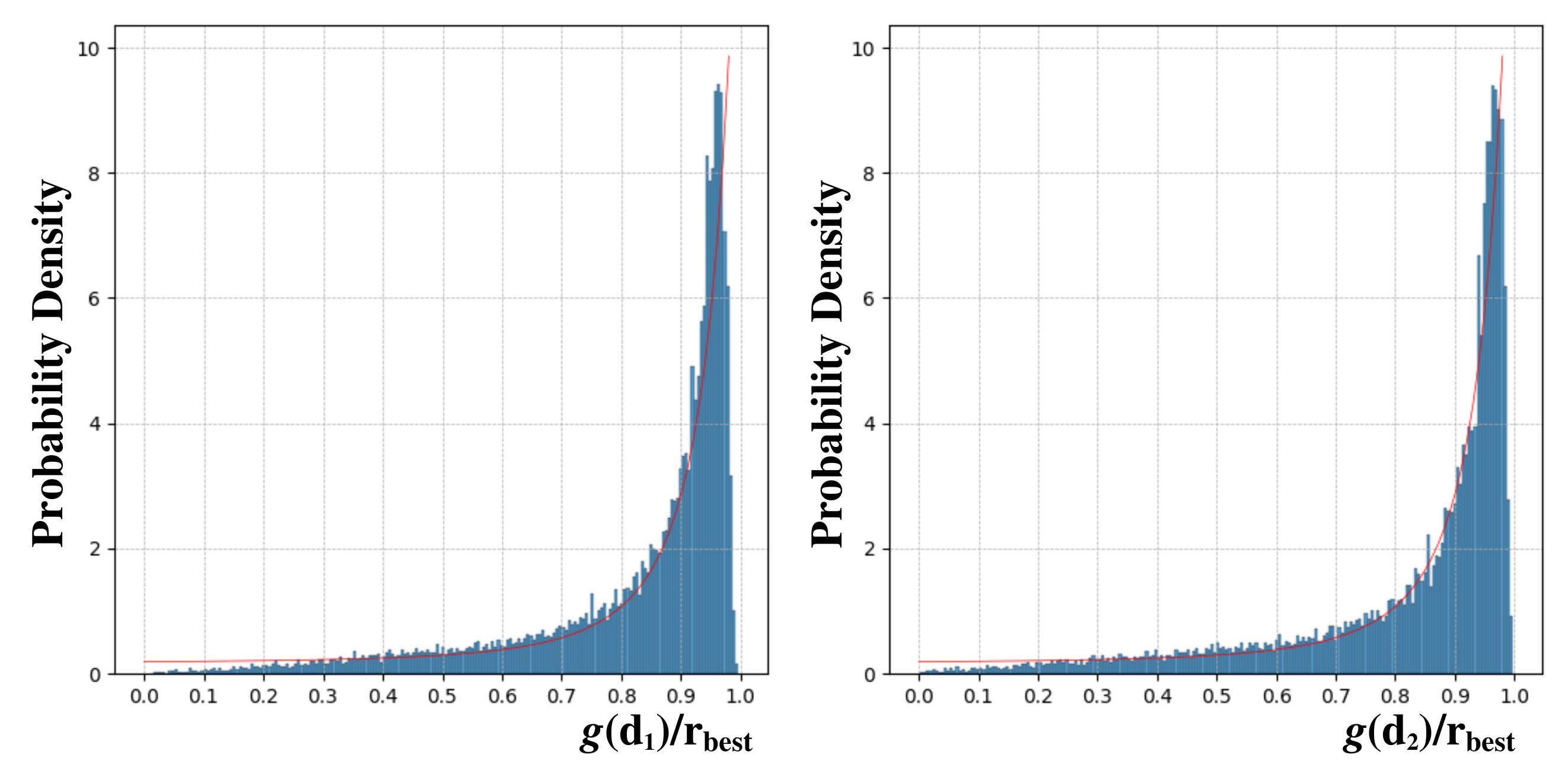}   
    \captionsetup{justification=raggedright, singlelinecheck=false}
    \caption{The distribution of decision boundaries \(g(\bm{d_1}) \text{ and } g(\bm{d_2})\)}
    \label{fig:Same_distribution}
\end{figure}

To ease analysis, we focus on the normalized distribution of \(g(\bm{d_1})/r_\text{best}\) and \(g(\bm{d_2})/r_\text{best}\) within the range \( [0, 1]\). We calculate and plot the distribution statistics of \(g(\bm{d_1})/r_\text{best}\) and \(g(\bm{d_2})/r_\text{best}\) using the \texttt{seaborn} package in Python, as shown in the blue histogram in Figure~\ref{fig:Same_distribution}. 
Apparently, the empirical distributions depicted in this figure are highly similar. Furthermore, we use the Python \texttt{scipy.stats} package to perform a Kolmogorov-Smirnov test on the samples of \(g(\bm{d_1})/r_\text{best}\) and \(g(\bm{d_2})/r_\text{best}\). The result shows that the p-value is less than 0.000001, indicating that \(g(\bm{d_1})/r_\text{best}\) and \(g(\bm{d_2})/r_\text{best}\) can be considered to follow the same distribution.

Furthermore, based on the shape of the probability density function, we decide to adopt the inverse proportional function \(\rho(r)=\frac{a}{(|-r/r_{\text{best}}+d|)^b+c}\) with \(r \in [0, r_{\text{best}}]\) to estimate the observed distribution of the decision boundaries, where \(a\), \(b\), \(c\), and \(d\) are the estimation parameters. We use the \texttt{scipy.optimize} package to determine the parameter values: \(a = 0.0313\), \(b = 3.066\), \(c = 0.168\), and \(d = 1.134\). The estimated density function \(\rho(r)\) is shown as the red curve in Figure~\ref{fig:Same_distribution}.

\begin{table}[htbp]
\centering
\captionsetup{justification=raggedright, singlelinecheck=false}
\caption{Comparison of ADBA, ADBA-md, and RayS on MNIST and CIFAR-10 dataset}
\label{table-general_ADBA_mnist_cifar}
\setlength{\tabcolsep}{15.0mm}
\begin{tabular}{l|cc}
\hline
\hspace{2em}\textbf{Datasets}&\textbf{CIFAR-10}&\textbf{MNIST}\\
 \textbf{Attacks   }&& \\  \hline
\textbf{RayS}&813.3&697.4 \\
 &(364)&(219) \\
 &99.8\%&\textbf{97.0\%} \\
 \textbf{ADBA}&607.2&600.6 \\
 &(216)&(192) \\
 &99.5\%&95.3\% \\ 
\textbf{ADBA-md}&\textbf{482.3}&\textbf{504.6} \\
 &\textbf{(192)}&\textbf{(166)} \\
 &\textbf{99.9\%}&\textbf{97.0\%} \\ \hline
\end{tabular}
\end{table}

\section{Generality and transferability of ADBA-md}
\label{appendix ADBA-md mnist cifar}
In Subsection~\ref{section optimize ADB}, ADBA-md uses \(\rho(r)\) to optimize ADB. In Appendix~\ref{appendix Same_distribution_function_rho}, the parameters for estimating \(\rho(r)\) are determined by analyzing the decision boundaries of many different perturbation directions obtained on the ResNet50 model using the ImageNet dataset.

Despite of the efforts spent in estimating \(\rho(r)\), we found empirically that the effectiveness of using \(\rho(r)\) is not sensitive to the estimation parameters. Hence, the practical use of ADBA-md is not limited to the ResNet50 model and the ImageNet dataset. ADBA-md can be directly applied to other datasets and models without re-estimating \(\rho(r)\).

In this appendix, we design additional experiments to demonstrate the general applicability of ADBA-md without re-estimating \(\rho(r)\). Particularly, we randomly select 1000 images from the CIFAR-10 dataset \cite{krizhevsky2009cifar} and attack a pre-trained 7-layer CNN model as mentioned in \cite{chen_rays_2020} based on the selected images. Additionally, on the MNIST dataset \cite{lecun1998mnist}, we randomly select 1000 images and attack a pre-trained 7-layer CNNs studied in \cite{chen_rays_2020}. We set the perturbation strength threshold \(\epsilon=0.031\) for CIFAR-10 and  \(\epsilon=0.15\) for MNIST. The query budget is set to 10000 for both models. The estimation parameters in the distribution function \(\rho(r)=\frac{a}{(|-r/r_{\text{best}}+d|)^b+c}\), \(r\in[0,r_{\text{best}}]\) are set to be \(a = 0.0313\), \(b = 3.066\), \(c = 0.168\), and \(d = 1.134\) according to Appendix~\ref{appendix Same_distribution_function_rho}. Meanwhile, the search tolerance \(\tau\) is set to be 0.002.

Table~\ref{table-general_ADBA_mnist_cifar} shows the attack success rate and the average (median) number of queries achieved by ADBA, ADBA-md, and RayS. Our results clearly indicate that, ADBA-md still maintains the highest attack success rate and the lowest number of queries (on average and in median). This confirms that the distribution function \(\rho(r)\) in ADBA-md is not sensitive to its estimation parameters. ADBA-md can be easily applied to a variety of target models and datasets without re-estimating \(\rho(r)\).

\end{document}